\documentclass[letterpaper, 10 pt, conference]{ieeeconf}  
\IEEEoverridecommandlockouts                              
\usepackage[english]{babel}
\usepackage[utf8]{inputenc}

\usepackage{cite}
\usepackage{amsmath,amssymb,amsfonts}
\usepackage{algorithmic}
\usepackage{graphicx}
\usepackage{textcomp}
\usepackage{xcolor}
\usepackage{multicol}
\usepackage{multirow}
\usepackage[export]{adjustbox}

\def\etal{\textit{et al.}}

\begin{document}

\title{\LARGE \bf Grounding Linguistic Commands to Navigable Regions\\
\thanks{\textsuperscript{1} KCIS, International Institute of Information Technology, Hyderabad}
\thanks{\textsuperscript{*} Equal contribution from the first three authors.}
}

\author{{Nivedita Rufus\textsuperscript{*1} }
\and
{ Kanishk Jain\textsuperscript{*1} }
\and
{ Unni Krishnan R Nair\textsuperscript{*1} }
\and
{Vineet Gandhi\textsuperscript{1} }
\and
{ K Madhava Krishna\textsuperscript{1}} 
} 

\maketitle

\begin{abstract}
Humans have a natural ability to effortlessly comprehend linguistic commands such as ``park next to the yellow sedan" and instinctively know which region of the road the vehicle should navigate. Extending this ability to autonomous vehicles is the next step towards creating fully autonomous agents that respond and act according to human commands. To this end, we propose the novel task of Referring Navigable Regions (RNR), i.e., grounding regions of interest for navigation based on the linguistic command. RNR is different from Referring Image Segmentation (RIS), which focuses on grounding an object referred to by the natural language expression instead of grounding a navigable region. For example, for a  command ``park next to the yellow sedan," RIS will aim to segment the referred sedan, and RNR aims to segment the suggested parking region on the road. We introduce a new dataset, Talk2Car-RegSeg, which extends the existing Talk2car~\cite{deruyttere2019talk2car} dataset with segmentation masks for the regions described by the linguistic commands. A separate test split with concise manoeuvre-oriented commands is provided to assess the practicality of our dataset. We benchmark the proposed dataset using a novel transformer-based architecture. We present extensive ablations and show superior performance over baselines on multiple evaluation metrics. A downstream path planner generating trajectories based on RNR outputs confirms the efficacy of the proposed framework.

\end{abstract}

\begin{figure}[t]
    \centering
    \begin{tabular}[t]{p{0.48\textwidth} }
    { \centering \small {\textit{\textbf{Command: }``Turn in the direction where the man is pointing.''}}}\\
        \includegraphics[width=0.48\textwidth]{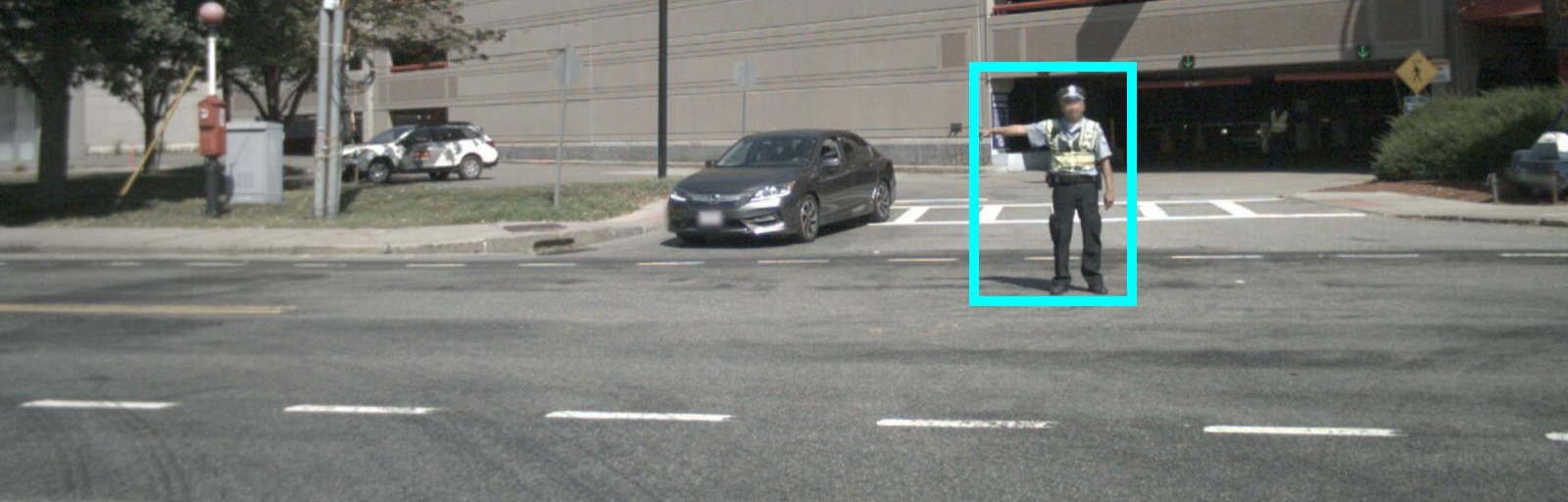} \\
        \includegraphics[width=0.48\textwidth]{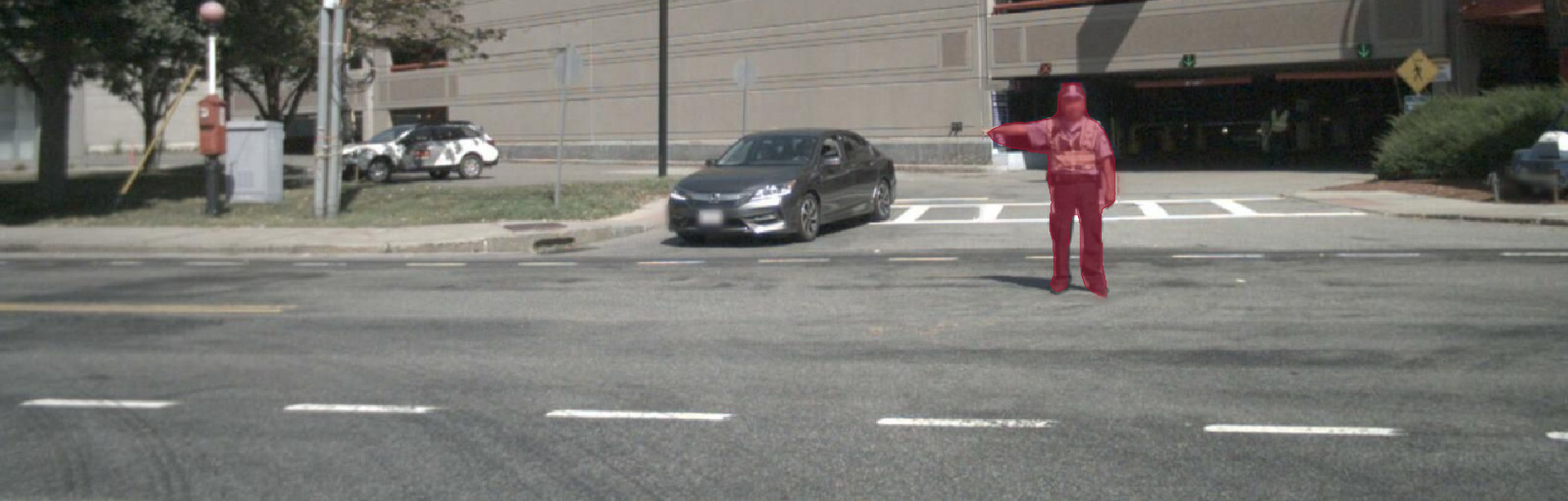}\\
        \includegraphics[width=0.48\textwidth]{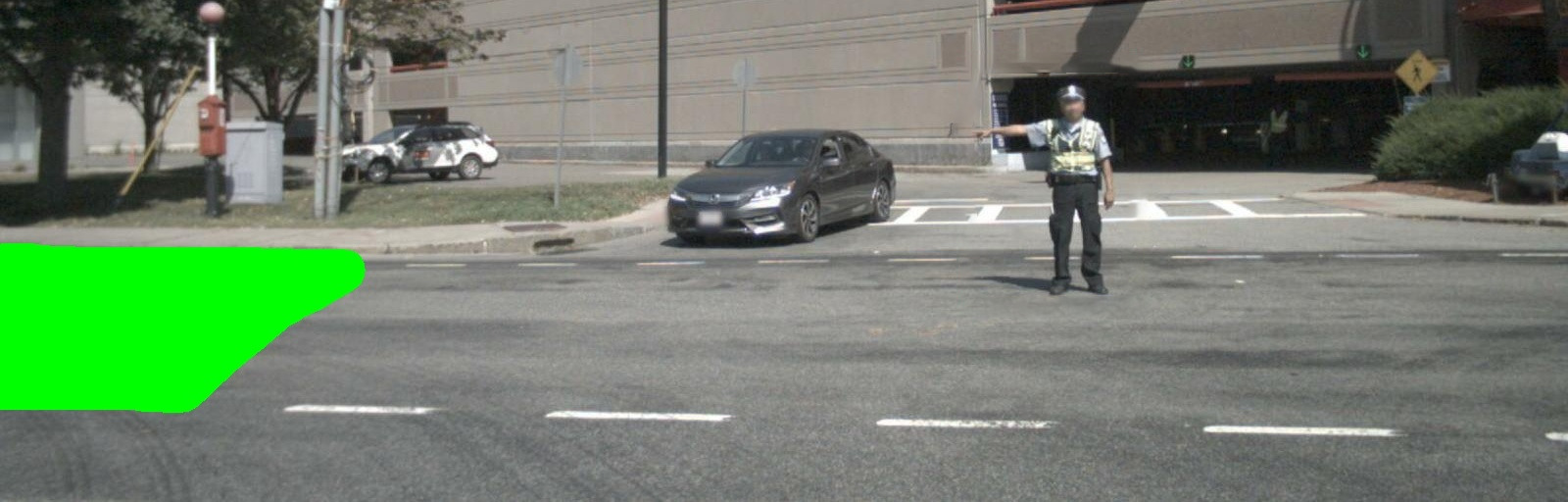}
    \end{tabular}
    \caption{Given a natural language command, REC (top image) predicts a bounding box (cyan box) around the referred object and RIS (the middle image) predicts a segmentation map around the referred object. In the context of an AD application, such predictions are not immediately amenable to downstream tasks like planning. E.g. predicting the man in the above example does not indicate where the car should go. In contrast, our work aims to directly predict regions on the road given a natural language command (green colour annotation, bottom image).}
    \label{fig:teaser}
\end{figure}
\section{Introduction}

Autonomous Driving (AD) is concerned with providing machines with driving capabilities without human intervention. Much of the existing work on autonomous driving has focused on modular pipelines, with each module specializing in a separate task like detection, localization, segmentation and tracking. Collectively, these tasks form the vehicle's active perception module, enabling it to perform driver-less navigation with some additional help from prior generated detailed high definition maps of the route. However, the current setup does not allow the capability to intervene and augment the vehicle's decision-making process. For example, post reaching the destination, the rider may want to give specific guidance on the place to park the car suiting his/her convenience, e.g. ``park between the yellow and the red car on the left".  

Similarly, sometimes the rider may wish to intervene to resolve ambiguities or to perform the desired action, e.g. ``the road appears to be blocked, please move to the left lane" or ``I see my friend walking on the left, please slow down and pick him up". In a chauffeur-driven car, the above scenarios are commonplace, as a human can easily understand the natural language commands and manoeuvre the car accordingly. In this work, we aim to extend similar abilities to a self-driving vehicle, i.e. the vehicle takes the natural language command and the current scene as input and predicts the region of interest where the car must navigate to execute the command. A downstream planner can take this region as input and predict the trajectory or set of manoeuvres to perform the desired navigation.

One of the fundamental tasks necessary to attain the above capabilities is comprehending the natural language command and localising it in the visual domain. The problem is formally known as visual grounding, and it has seen a surge of interest in the recent past. The interest is primarily driven by the success of deep learning models in computer vision and natural language processing. Most of the current literature in visual grounding focuses on localising an object of interest. The object of interest can be grounded either using bounding boxes (Referring Expression Comprehension) or using segmentation masks (Referring Image Segmentation). We focus on the latter type of grounding, but instead of grounding objects, we ground regions of interest on the road. Grounding regions of interest are more natural from a navigation point of view for self-driving vehicles than the grounding of objects. Even if the referred object is correctly grounded, it leaves ambiguity on where to take the vehicle. In contrast, the task of Referring Navigable Regions proposed in our work provides feasible areas as a goal point. A motivating example is illustrated in Figure~\ref{fig:teaser}.



To this end, we introduce a novel multi-modal task of Referring Navigable Regions (RNR), intending to ground navigable regions on the road based on natural language command in the vehicle's front camera view. Compared to RIS task, RNR task involves two-level understanding of the scene. In the first level, the referring object has to be identified and in second level the appropriate region for navigation has to be identified based on the referred object. For instance, consider the command ``park beside the white car near the tree", in addition to locating the ``white car" near the ``tree", the RNR task also has to predict an appropriate region where the command can be executed. Consequently, we propose a new dataset, Talk2Car-RegSeg, for the proposed task. This dataset is built on top of the existing Talk2car~\cite{deruyttere2019talk2car} dataset. In addition to the existing image-command pairs, we provide segmentation masks for the regions on the road where the vehicle could navigate to execute the command. We benchmark the proposed dataset with a transformer-based grounding model that can capture correlations between visual and linguistic features through the self-attention mechanism. We compare the proposed model against a set of baselines and present thorough ablation studies. We highlight the proposed task's practicability through a downstream planning module that computes a navigation trajectory to the grounded region. To summarize, the main contributions of this paper are the following:



\begin{itemize}
\item We introduce the novel task of RNR for applications in autonomous navigation. 
\item We present a new Dataset, Talk2Car-RegSeg, for this task. Here, we augment the existing Talk2car dataset with segmentation masks for navigable regions corresponding to the command.
\item We benchmark the dataset using a novel transformer based model and a set of baseline approaches. We present thorough ablations and analysis studies on the proposed dataset (e.g. action type of commands, the length of commands) to assess its applicability in realistic scenarios.

\end{itemize}


\section{Related Work}

\begin{figure*}[t]
\centering
\includegraphics[width=1.0\textwidth]{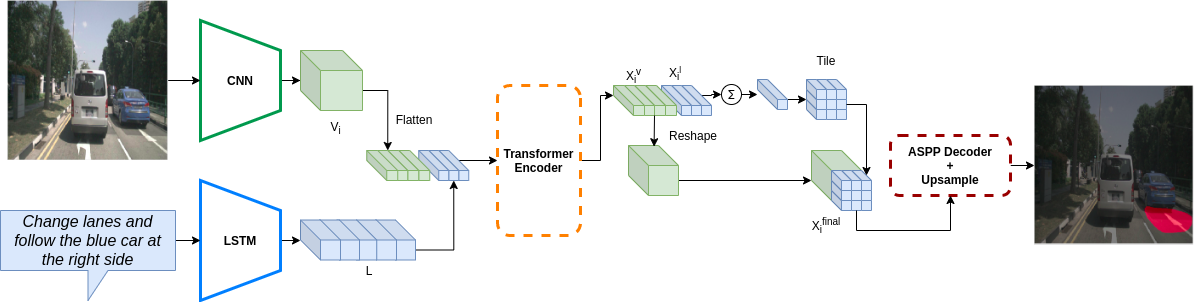}
\caption{Network architecture for the Transformer Based Model (TBM).}
\label{pipeline}
\end{figure*}

\subsection{Referring Expression Comprehension}
Referring Expression Comprehension (REC) predicts a rectangular bounding box in an image corresponding to the input phrase or the sentence. While object detection~\cite{ren2015faster,redmon2016you} predicts bounding boxes for a pre-defined set of categories, REC does not limit on a category list. Nonetheless, the task of REC does take inspiration from the object detection pipeline. In the most commonly used framework, a set of bounding box region proposals are first generated and then evaluated against the input sentence~\cite{plummer2015flickr30k, rohrbach2016grounding}. In the robotics community, significant progress has been made on using REC in Human-Computer Interactions~\cite{shridhar2017grounding, shridhar2020ingress}. REC has also been explored on autonomous driving applications, following the introduction of the Talk2Car dataset~\cite{deruyttere2019talk2car}. Rufus~\etal~\cite{rufus2020cosine} use softmax on cosine similarity between region-phrase pairs and employ a cross-entropy loss. Ou~\etal~\cite{ou2020attention} employ multimodal attention using individual keywords and regions.  Despite significant progress in REC, bounding box based localization is not accurate enough to capture the shape of the referred object and struggle with objects at a small scale. Furthermore, just predicting the bounding box is insufficient for the task of navigation (as illustrated in Figure~\ref{fig:teaser}).

\subsection{Referring Image Segmentation}
 Referring Image Segmentation (RIS) task was introduced in \cite{hu2016segmentation} to alleviate the problems associated with REC by predicting a pixel-level segmentation mask for the referring object based on the referring expression. \cite{liu2017recurrent} propose convolutional multimodal LSTM to encode the sequential interactions between individual words and pixel-level visual information. \cite{Shi2018KeyWordAwareNF} utilize query attention and key-word-aware visual context to model relationships among different image regions, according to the corresponding query. More recent works, \cite{hui2020linguistic} model multimodal context by cross-modal interaction and guided through a dependency tree structure, \cite{huang2020referring} progressively exploits various types of words in the expression to segment the referent in a graph-based structure. In contrast to existing works on RIS that directly refer to objects in an image, we ground the region adjacent to the object to provide navigational guidance to a self-driving vehicle. To the best of our knowledge, our work is the first paper to explore the referring image segmentation in the context of autonomous driving and propose the task of Referring Navigable Region.


\subsection{Language Based Navigation}
Most of the literature on language-based navigation has focused on indoor navigation~\cite{shah2018follownet, zang-etal-2018-translating, wang2019reinforced}. Typically the input to these approaches is a longer text (a paragraph), and the goal is to reach the required destination in an indoor 3D environment (long trajectory prediction). Shah~\etal\cite{shah2018follownet} utilized attention over linguistic instructions conditioned on the multi-modal sensory observations to focus on the relevant parts of the command during navigation task. \cite{zang-etal-2018-translating} approach the language-based navigation task as a sequence prediction problem. They translate navigation instructions into a sequence of behaviours that a robot can execute to reach the desired destination. Wang~\etal~\cite{wang2019reinforced} enforces cross-modal grounding both locally and globally via reinforcement learning. 

Sriram~\etal~\cite{Sriram-IROS-2019} attempt language-based navigation in an autonomous driving scenario. They generate trajectory based on natural language command by predicting local waypoints. However, their work limits to eight specific behaviours like \textit{take left, take right, not left}, etc. The only object considered in their work is a traffic signal. Our work considers much richer language instructions encompassing many objects. Furthermore, RNR predicts a segmentation map instead of a single local waypoint or a trajectory corresponding to a set of sentences. Segmentation masks unlike single waypoint encourage multiple trajectory possibilities and options to navigate into that region for a downstream planning or navigation task.



\section{Dataset}

%
The proposed Talk2Car-RegSeg dataset is built on top of the Talk2Car dataset, an object referral dataset containing commands written in natural language for self-driving cars.  
The original dataset had textual command with a specific action, referring to an object in the image, and the object of interest was referred to using a bounding box. However, for AD applications, as referring directly to objects is not amenable for downstream tasks like planning, we augmented the original dataset with segmentation masks corresponding to navigable regions. The newly created Talk2Car-RegSeg dataset has 8349 training and 1163 validation image-command pairs, similar to those used in the original dataset. We observed that the commands in Talk2Car's validation set are very complex as they are verbose, and in a significant number of cases, there were more than one actions in a single command ex: "we need to turn right instead of left, as soon as this truck pulls forward, move over to the right lane behind it." We first present results on the full validation set; however, to evaluate the performance in a controlled setting, we also curated a novel test split (Test-RegSeg). Test-RegSeg contains 500 randomly selected images from the validation set with newly created commands. The commands in the Test-RegSeg split are simplified and straightforward. We present results, baseline comparisons, and ablations on both the complex instruction validation set and the curated simpler instruction set (Test-RegSeg). In the rest of the paper, we consider Test-RegSeg as our test set. The dataset and the code-base will be  released at (rnr-t2c.github.io). In the next section, we describe the dataset creation process.


\subsection{Dataset Curation}

The authors of the paper manually annotated the navigable regions in each image based on the linguistic command. A simple Graphical User Interface (GUI) was created to make the annotation process straightforward. In the GUI, each annotator sees the image, the linguistic command, and the bounding box for the referred object in the scene. We used ground truth bounding boxes from the original Talk2Car dataset as a reference to identify the referred object in the scene to resolve ambiguities and only focus on annotating regions of interest. 


To verify the quality of annotations, we hired a group of three students from the institute for the role of annotation reviewers. All the reviewers were briefed on the task and were asked to ensure that all feasible regions for navigation were annotated in the image. Depending on the reviewers' assessment, each annotation could be either accepted or sent for re-annotation. An annotation was accepted if at least two reviewers concurred on it. In the other cases, images were sent for re-annotation with reviewer comments for annotation refinement. This process was repeated iteratively until all annotations were qualitatively and logically acceptable. 

Since navigation is a flexible activity in terms of different ways of performing it, we involved multiple people as annotators and reviewers to capture different perspectives and incorporate those in our dataset.

\section{Methodology}

Given an image $I$ from a front-facing camera on the autonomous vehicle and a natural language command $Q$, the goal is to predict the segmentation mask of the region in the image where the vehicle should navigate to fulfil the command. Here, the command $Q$ corresponds to a navigable action in the image. Compared to the traditional task of Referring Image Segmentation, the proposed task is more involved as the ground truth masks are unstructured. To correctly identify the regions of interest, the model should be able to learn correlations between words in commands and regions in the image. We propose two models for this task, a baseline model and another transformer-based model. The feature extraction process is the same for both models. They only differ in multi-modal fusion and context modelling. We describe the feature extraction process in the next section and describe each model in the subsequent sections.

\subsection{Feature Extraction} 
We extract visual features from image using a DeepLabV3+ \cite{chen2017deeplab} backbone pre-trained on semantic segmentation task. Since hierarchical features are beneficial for semantic segmentation, we derive hierarchical features $V_i$ of size $C_i$$\times$$H_i$$\times$$W_i$ with $i \in \{ 2,3,4\}$, corresponding to last 3 layers, namely $layer2$, $layer3$ and $layer4$ of CNN backbone. Here $H_i, W_i$ and $C_i$ correspond to height, width and channel dimension of visual features corresponding to each level. Each $V_i$'s are transformed to same spatial resolution $H_i = H$, $W_i = W$ and channel dimension $C_i = C_v$ using $3 \times 3$ convolutional layers. We initialize each word in the linguistic command with the GloVe word embedding, which are then passed as input to LSTM encoder, to generate linguistic feature for the command. We denote the linguistic feature as $ L = \{l_1, l_2, ..., l_T\} $, where $T$ is the number of words in the command and $l_i \in \mathbb R^{C_l}$, $i \in \{1,2,..T\}$ is the linguistic feature for the i-th word. In all our experiments, $C_v = C_l = C$ and $H = W = 14$.

\subsection{Baseline Model}
\label{subsection:baseline}

Our baseline model is inspired from \cite{hu2016segmentation}, we first compute the command feature $L_{avg} \in \mathbb R^{C_l} $ by averaging all the word features $l_i$ in $L$. In order to fuse visual features with linguistic features, we repeat the command feature $L_{avg}$ along each spatial location in the visual feature map and then concatenate the features from both modalities along channel dimension to get a multi-modal feature $M_i$ of shape $\mathbb R^{(C_v + C_l) \times H \times W}$. Since the number of channels, $C_v + C_l$ can be large, we apply $1 \times 1$ convolution to $M_i$ reduce the channel dimension to $C$, resulting in final multi-modal feature $X^{final}_i \in \mathbb R^{C \times H \times W}$.

\subsection{Transformer Based Model}
Our baseline model has few shortcomings: $(1)$ the word-level information is lost when all word features are averaged to get the command feature. $(2)$ multi-modal context is not captured effectively with a concatenation of visual and linguistic features. To address these shortcomings, we propose a transformer-based model (TBM). We borrow from the architecture of DETR \cite{carion2020endtoend} for our transformer based model. Specifically, we adopt their transformer encoder, and along with image features $V_i$, we also pass textual feature $L$ as input by concatenating features from both modalities along length dimension, resulting in multi-modal feature $M_i$ of shape $\mathbb R^{C \times (HW + T)}$, $T$ is the number of words in the input command. $M_i$ is passed as input to the transformer encoder, where self-attention enables cross-modal interaction between word-level and pixel-level features, resulting in multi-modal contextual feature $X_i$ of the same shape as $M_i$. Since all word features are utilized during the computation of $X_i$, the word -level information is preserved, and because of inter-modal and intra-modal interactions in the transformer encoder, the multi-modal context is captured effectively. To predict a segmentation mask from $X_i$, we need to reshape it to the same spatial resolution as $V_i$, i.e., $H \times W$. So, $X_i$ is separated into attended visual features, $X^{v}_{i}$  and attended linguistic features, $X^{l}_{i}$ of dimensions $\mathbb R^{HW \times C}$ and $\mathbb R^{T \times C}$, respectively. $X^{l}_{i}$ is averaged across length dimension and concatenated with $X^{v}_{i}$ along the channel dimension and reshaped to result in a feature vector of shape $\mathbb R^{2C \times H \times W}$. Finally, $1 \times 1$ convolution is applied to give final multi-modal feature $X^{final}_i \in \mathbb R^{C \times H \times W}$.

\subsection{Mask Generation}
To generate the final segmentation mask, we stack $X^{final}_i$ for all levels and pass them through Atrous Spatial Pyramid Pooling (ASPP) Decoder from \cite{chen2018encoderdecoder}. We use $3\times 3$ convolution kernels followed by bi-linear upsampling to predict the segmentation mask at a higher resolution. Finally, sigmoid non-linearity is applied to generate pixel-wise labels for segmentation mask $Y$.
Both baseline and transformer-based models are trained end-to-end using binary cross-entropy loss between predicted segmentation mask $Y$, and the ground truth segmentation mask $G$.


\section{Experiments}

\textbf{Implementation Details}: We use DeepLabV3+ \cite{chen2017deeplab} with ResNet-101 as backbone for visual feature extraction. Our backbone is pre-trained on the Pascal VOC-12 dataset with the semantic segmentation task. Input images are resized to $448\times 448$ spatial resolution. We use 300d GloVe embeddings pre-trained on Common Crawl 840B tokens~\cite{pennington-etal-2014-glove}. The maximum length of commands is set to $T=40$ and for both visual and linguistic features, channel dimension $C = 512$. Batch size is set to $64$, and our models are trained using AdamW optimizer with weight decay of $5e^{-4}$, the initial learning rate is set to $1e^{-4}$ and gradually decreased using polynomial decay by a factor $0.5$. 

\textbf{Evaluation Metrics}: In the proposed dataset, the ground truth segmentation masks incorporate all viable regions of interest for navigation, so any point inside the annotated region can be used as a target destination. Considering this aspect of our dataset, we evaluate our models' performance on three metrics, namely, Pointing Game, Recall@$k$ and Overall IOU. Pointing Game Metric (PGM) indicates the per cent of examples where the highest activated point lies inside the ground truth mask. It is calculated in the following way:
\begin{equation}
PGM\ Score = \frac{\#\ of\ hits}{total\ examples}
\end{equation}
A $hit$ occurs when the highest activated pixel lies inside the ground truth segmentation mask. It is possible that in some cases, the point with the highest activation is slightly outside the annotated ground truth region. However, the overall prediction is almost correct. Recall@$k$ metric is used to underscore the performance of models in such scenarios. Recall@$k$ metric is calculated as the proportion of examples where at least one of the top-$k$ points lies inside the ground truth mask. Finally, we also show results with the Overall IOU metic. Previous works commonly use the Overall IOU metric \cite{hu2016segmentation, hui2020linguistic, huang2020referring} for RIS task, it is calculated as the ratio of total intersection and total union between the predicted and ground truth segmentation masks across all examples in the dataset.


\begin{table*}[t!]
\centering
    \caption{recall$@k$ metrics for the validation and test set}
        \begin{tabular}{|p{1.2cm}|c|c|c|c|c|c|c|c|c|c|c|c|}
    \hline
    \multicolumn{1}{|c|}{\multirow{3}{*}{\textbf{Method}}}&
    \multicolumn{12}{c|}{\textbf{Recall @k for PGM}}\\ \cline{2-13}
    & \multicolumn{2}{c|}{$k=5$}& \multicolumn{2}{c|}{$k=10$}& \multicolumn{2}{c|}{$k=50$}& \multicolumn{2}{c|}{$k=100$}& \multicolumn{2}{c|}{$k=500$}& \multicolumn{2}{c|}{$k=1000$} \\ \cline{2-13}
    & Val set& Test set& Val set& Test set& Val set& Test set& Val set& Test set& Val set& Test set& Val set& Test set\\ \hline \hline
    Baseline& 51.84 & 69.80 & 52.71 & 69.80 & 55.29 & 72.20 & 56.92 & 73.80 & 64.49 & 79.60 & 69.64 & 83.80 \\\hline
    TBM& 59.67 & 77.00 & 60.53 & 78.20 & 63.19 & 79.60 & 64.91 & 81.80 & 72.65 & 86.60 & 78.07 & 90.20 \\\hline
    \end{tabular}
    \label{table:Recall_at_k}
\end{table*}

\subsection{Experimental Results}
In this section, we present the experimental results on different evaluation metrics. For all metrics, we compute the results on both validation and test split.

\textbf{Pointing Game}: Results on pointing game metric are presented in Table \ref{table:pointing_game}. First, we compare against a centre baseline to showcase the diversity of localization of annotated regions and ensuring that our dataset is free of centre-bias. In this baseline, the image's centre point is considered as the point with the highest activation for the pointing game metric. PGM score is $5.07$\%, and $6.61$\% for this baseline on validation and test splits, respectively, thus clearing our dataset for centre-bias. Next, we compare against the baseline model presented in Section \ref{subsection:baseline}. Our baseline model gives a PGM score of $49.78$\% and $66.80$\% on validation and test split, respectively. The test split score is high as the commands for the test split are simple and concise compared to those in the validation split. The Transformer-Based Model (TBM) gives higher numbers than both the baselines on both splits. We observe an improvement of  $8$\% and $10$\% over the baseline model for validation and test split, respectively. This improvement indicates the benefits of using the proposed multi-modal attention in the transformer-based approach, which can effectively model word-region interactions.



\begin{table}[t!]
    \centering
    \caption{pgm and overall iou for the validation and test set}
    \begin{tabular}{|p{1.2cm}|c| c| c| c|}
    \hline
    \multicolumn{1}{|c|}{\multirow{3}{*}{\textbf{Method}}}&
    \multicolumn{2}{c|}{\textbf{PGM}}& \multicolumn{2}{c|}{\textbf{Overall IOU}}\\ \cline{2-5}
    & Val set& Test set& Val set& Test set\\ \hline
    Baseline& 49.78& 66.80& 19.88& 29.28\\ \hline
    TBM& 58.03& 76.60& 22.17& 30.61\\ \hline
    \end{tabular}
    \label{table:pointing_game}
\end{table}

\begin{table}[t]
    \centering
    \caption{pgm for the validation data wrt command length where $T =$ number of words in a command}
     \begin{tabular}{|p{1.2cm}|c|c|c|}
    \hline
     \multicolumn{1}{|c|}{\multirow{2}{*}{\textbf{Method}}} & \multicolumn{3}{c|}{\textbf{PGM score on the Val set}}\\ \cline{2-4}
    & $T<10$ &  $10 \leq T<20$& $T \geq 20$\\ \hline
    Baseline& 52.09& 48.55& 44.00\\ \hline
    TBM& 60.00& 57.06& 52.00\\ \hline
    \end{tabular}
    \label{cmd-len}
\end{table}

\begin{table}[t]
    \centering
    \caption{pgm on the test set with commands for various maneuvers}
     \begin{tabular}{|p{0.9cm}|p{0.7cm}|p{0.7cm}|p{0.6cm}|p{1cm}|p{1cm}|p{.7cm}|}
    \hline
      \multicolumn{1}{|c|}{\multirow{2}{*}{\textbf{Method}}} & \multicolumn{6}{c|}{\textbf{PGM score on the test set}}\\ \cline{2-7}
    &Stop/ Park &Follow &Turn &Maintain Course &Go Slow/Fast &Change lanes\\ \hline
    Baseline& 62.83& 72.91& 74.19& 50.00& 76.31& 68.74\\ \hline
    TBM& 75.12& 88.23& 86.59& 83.34& 77.15& 84.62\\ \hline
    \end{tabular}
    
    \label{type-cmd}
\end{table}

\textbf{Recall@$k$}: Since our model mostly predicts connected and contiguous segmentation masks, Recall@$k$ metric indicates if we can approximately locate the correct area (where the highest activation point is near the ground truth region). Results for this metric are tabulated in Table \ref{table:Recall_at_k}, we consider values of $k$=\{5, 10, 50, 100, 500, 1000\}. Recall@$1$ is the same as pointing game metric as in both cases, we pick the point with the highest activation. As expected, the metric performance increases with the value of $k$. For transformer-based model, at $k=1000$, metric score is $78.07$\% and $90.20$\% for validation and test splits, respectively. 1000 pixels account for $\sim 0.5$\% of the overall pixels at the considered resolution. Hence, Recall@$1000$ metric suggests that we can approximately locate the correct area $90.20$\% of the time when using simpler and straightforward commands. This demonstrates the effectiveness of our approach and how we are able to reduce the search space for feasible regions for navigation significantly.

\textbf{Overall IOU}: Since any point inside the annotated region can be considered as a target destination, computing the overall IOU metric that is normally used in segmentation literature cannot serve as an adequate performance measure and is only an indicative measure. For example. if there are three parking slots available, even if the model predicts one of them, the prediction is correct, however, the IOU might be low. The results presented in Table \ref{table:pointing_game} illustrate this aspect. For the transformer-based model, the IOU metric is $22.17$\% for validation split and $30.61$\% for the test split. The numbers for test split are significantly better than those in validation split because of the simplicity of commands in test split. This metric illustrates the differences between RNR and RIS task and shows that the same metric cannot be used to judge the performance across these tasks.

\begin{figure*}[t]
        \centering
        \begin{tabular}[b]{p{0.23\textwidth} p{0.23\textwidth} p{0.23\textwidth} p{0.23\textwidth}}
        \includegraphics[width=\linewidth]{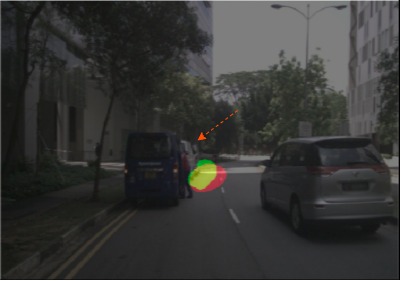} & 
        \includegraphics[width=\linewidth]{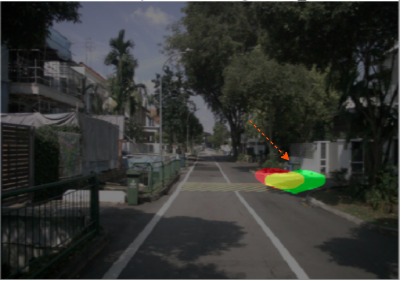} & 
        \includegraphics[width=\linewidth]{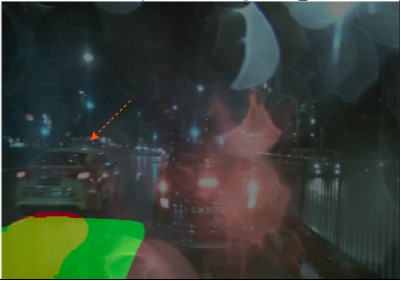} &
        \includegraphics[width=\linewidth]{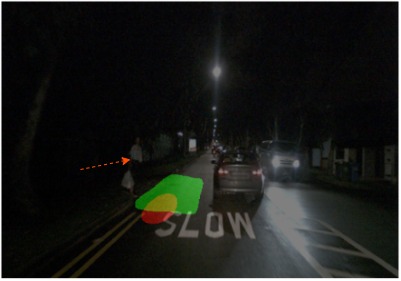} \\ 
        \centering{\scriptsize\textit{``Pull over by the white van"}} &
        \centering{\scriptsize\textit{``Park on the right by the green bin"}} &
        \centering{\scriptsize\textit{``Follow the yellow car that is in the left lane"}} &
        \centering{\scriptsize\textit{``pick up the woman on your left"}} \\ 
        \end{tabular}
        \caption{\centering{Qualitative Results for Successful Groundings. Our TBM network is able to ground the appropriate regions even in cases where the referred objects are barely visible. Red arrow is used to indicate the location of these referred objects.}}
       \label{fig:success}
\end{figure*}

\begin{figure*}[t]
    \centering
    \begin{tabular}[b]{p{0.23\textwidth} p{0.23\textwidth} p{0.23\textwidth} p{0.23\textwidth}}

        \includegraphics[width=\linewidth]{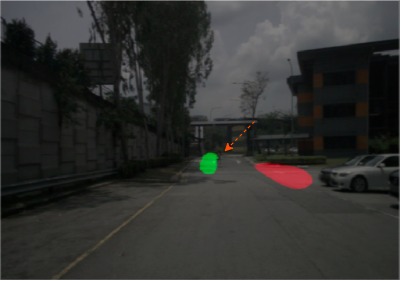}& 
        \includegraphics[width=\linewidth]{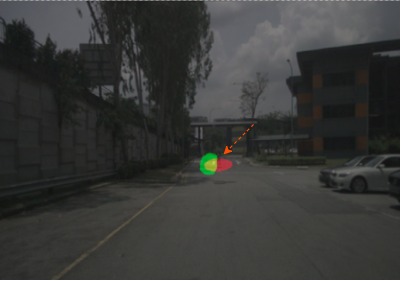}&
        \includegraphics[width=\linewidth]{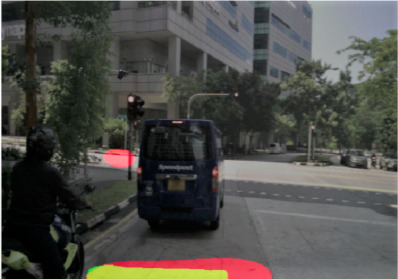}& 
        \includegraphics[width=\linewidth]{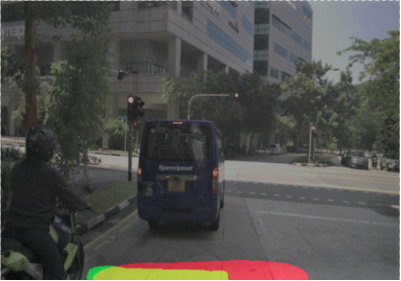} \\
         
         \centering{\scriptsize\textit{``Do not turn right as I said, carry on straight so I can talk to that person"}} &
         \centering{\scriptsize\textit{``Continue straight so I can talk to that person".}} &
         \centering{\scriptsize\textit{``Wait for the moped to continue before we turn left"}} &
         \centering{\scriptsize\textit{``Wait for the moped to leave"}} \\
    \end{tabular}
    \caption{\centering{Differences between the network performance on the original Validation set and the newly created Test split. For each image pair, example on the left is from the Validation split and one on the right is from the Test split with simplified commands. The ``person" in left pair of images is indicated using a red arrow.}}
    \label{fig:simple_split}
\end{figure*}

\subsection{Ablation Studies}\label{ablations}


In this section, we elaborate on the ablation and analysis studies performed on the proposed dataset and transformer-based model. We study various aspects of linguistic commands in the proposed Talk2car-RefSeg dataset on model performance. Specifically, we analyse the grounding performance of our model based on (1) the length of command and (2) the action specified in the command. As the commands in the test split are shorter than those in validation split, we conduct experiment (2) on the test split. Whereas based on the verbose nature of commands in the validation split, experiment (1) is conducted on the validation split. We used both baseline and transformer-based models and the pointing game metric for all the ablation studies.

\textbf{Based on Command Length}: We categorise the commands based on their length and present the ablation experiments in Table \ref{cmd-len}. All commands are grouped into three buckets, \{0-10, 10-20, $\geq$20\} based on their length. We observed that as the command length increases, the performance on the pointing game metric decreases. The performance gap between the first two buckets is $\sim$3\% , and that between the last two buckets is $\sim$5\% in TBM.  Since the commands in the validation split are long and complex, the network faces difficulties in grounding navigable regions for them. Some of the original talk2car dataset's commands contain unnecessary information from a grounding perspective, like addressing people using proper nouns. Because of this reason, we proposed a separate test split with concise commands. Length based grouping of commands in test split is not possible as the majority ($\sim$78\%) of them are less than 10 words long.

\textbf{Based on Action type}: Next, we classify each command to fixed basic action/manoeuvre categories and present the results on the pointing game metric in Table \ref{type-cmd}. For ``lane change" and ``turning" type of commands, our network can correctly predict the navigable region with high accuracy of $84.62$\% and $86.59$\%, respectively. For ``parking" based commands, we get a pointing game score of $75.12$\%. Parking is a challenging action to evaluate based on the Pointing game metric. In our dataset, the annotation mask is often relatively small for these cases, especially so when referring to a far away parking slot. The highest performance is observed on ``follow" type commands, where the metric is $88.23$\%. Commands with ``follow" action is easier to ground as in most cases, the navigable region is just behind the referred object (hence are less ambiguous). Results on these basic action/manoeuvre specific commands indicate the generality and practicality of our approach in realistic scenarios. 

%

\subsection{Qualitative Results}

In this section, we present the Qualitative results of our transformer-based approach. For all the example images in this section, Green, Red and Yellow signify the ground truth mask, the predicted mask and their intersection, respectively. Success cases of our approach are demonstrated in Figure \ref{fig:success}. The model successfully correlates textual words with regions in the image, ex: in the leftmost image, the model can successfully ground the region beside the white van, which is barely visible. Similarly, in the second image from the left, there are two green bins, the model can successfully resolve the ambiguity. The last two images demonstrate the performance of our model during night-time. In these cases, the referred object is barely visible, but the model can still infer the correct region.

Next, we showcase the differences between the original validation split and the newly created test split in Figure \ref{fig:simple_split}. For the leftmost image, the command is a bit confusing as there is a subtle negation involved. In order to resolve these issues, the model should be trained on training data with a large number of such instances. However, with the simplified command, the model predicts the correct region.  Similarly, for the second image, the model gets confused when there are multiple action words in the linguistic command. In this particular case, the model correctly predicted two regions corresponding to both actions ``wait" and ``turn left" despite the ambiguity in the command. This underscores our network's capability in effectively modeling the word-region interactions. After simplifying the command to include only one action ``wait", the model correctly predicted the corresponding navigable region.

\begin{figure}[t]
    \centering
    \begin{tabular}[b]{ p{0.23\textwidth} p{0.23\textwidth}}
         \includegraphics[width=\linewidth]{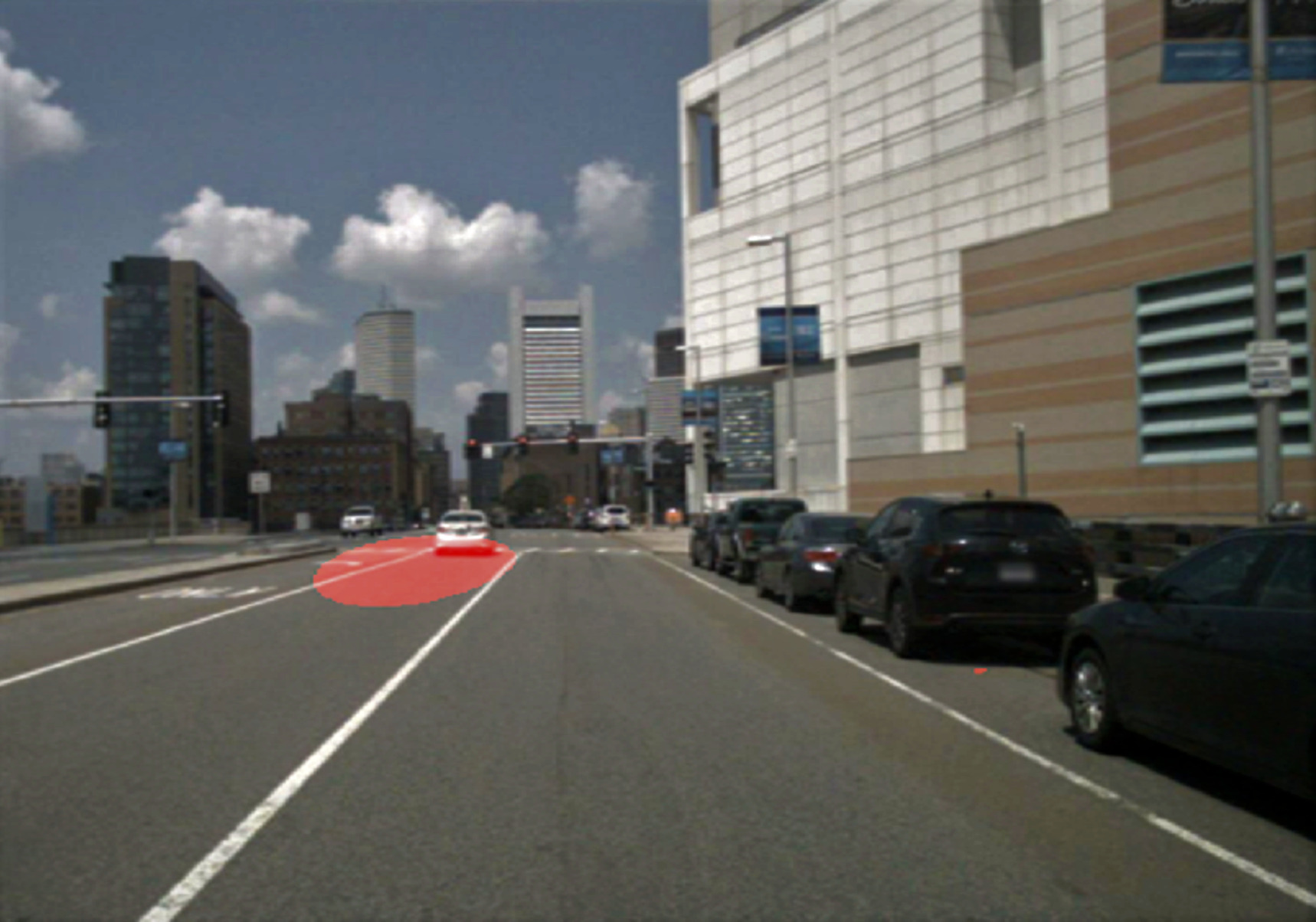}& 
        \includegraphics[width=\linewidth]{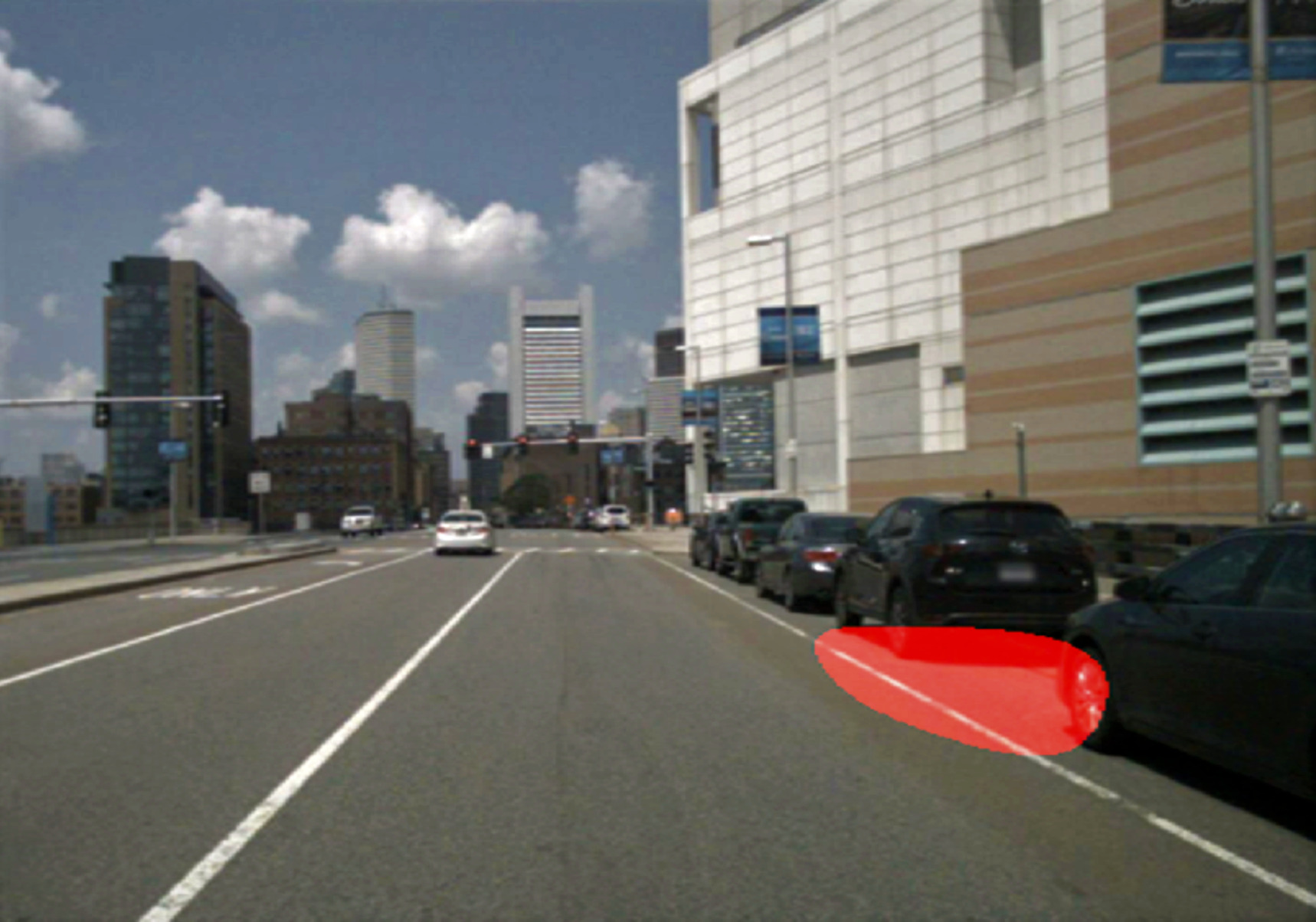}\\ 
         \centering{\scriptsize\textit{``Get into the next lane behind the car"}} &  
         {\scriptsize\textit{``Park in between the black cars on right"}} \\
        \includegraphics[width=\linewidth]{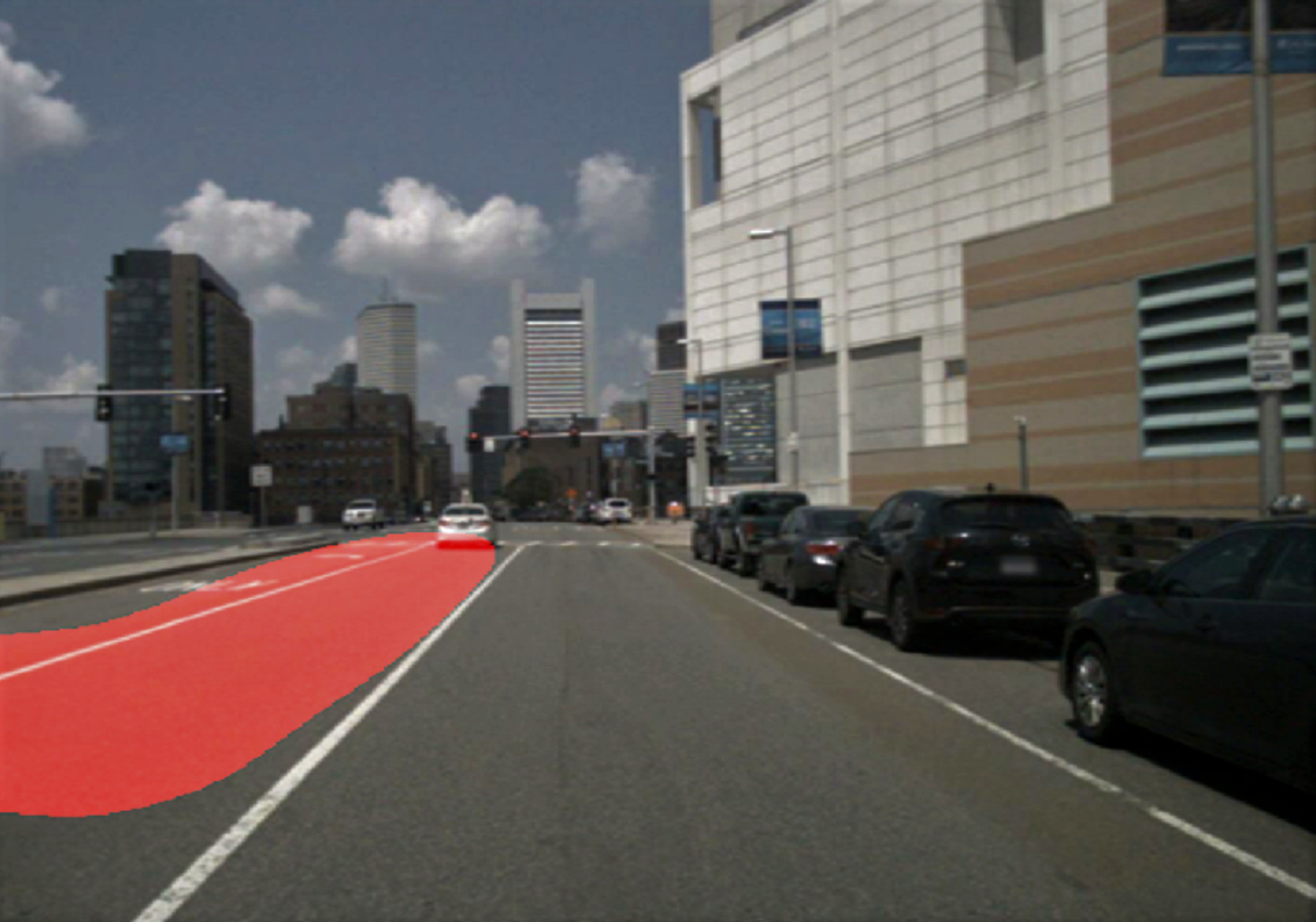}& 
        \includegraphics[width=\linewidth]{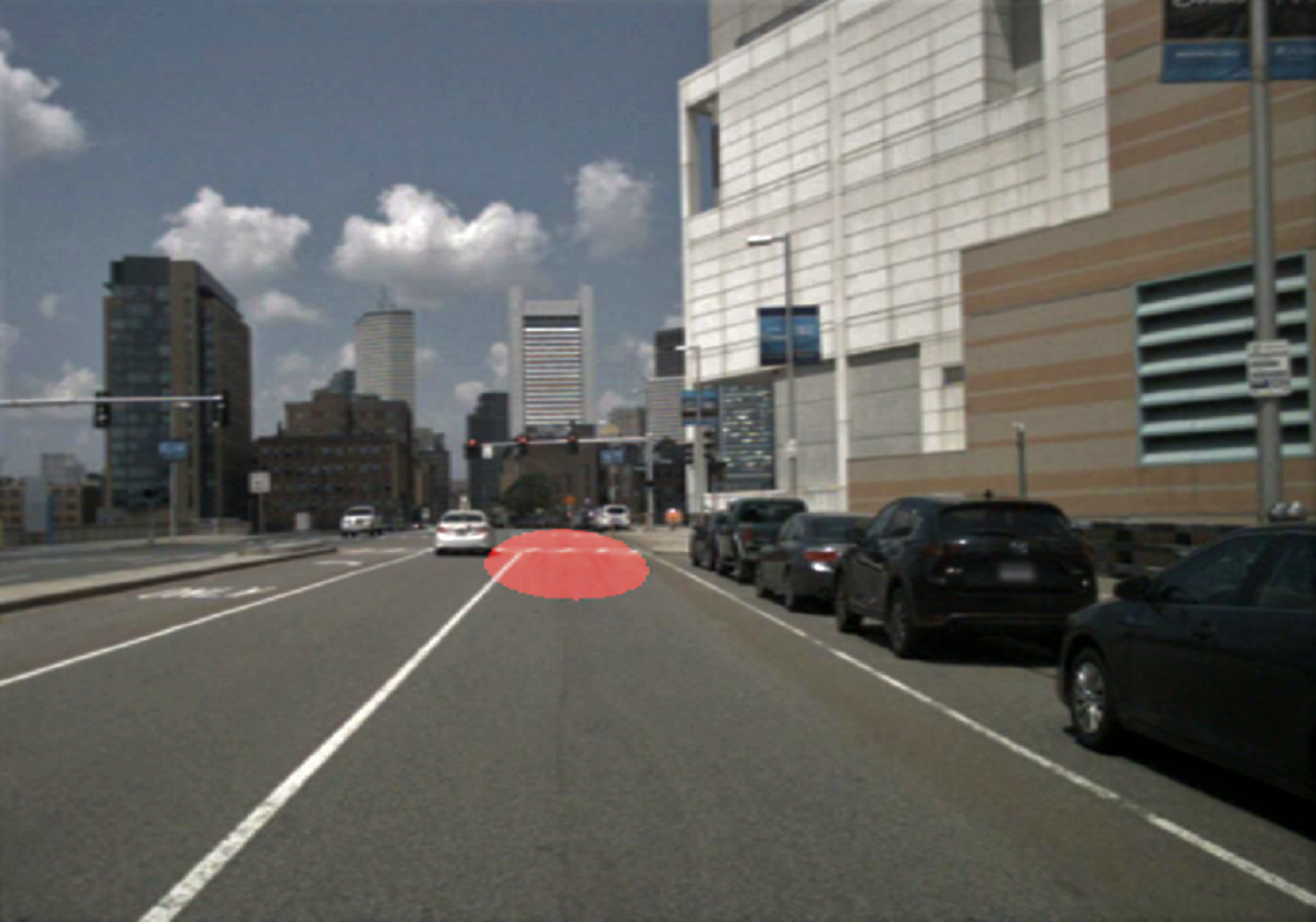} \\
         \centering{\scriptsize \textit{``Switch to the middle lane"}} &  
         \centering{\scriptsize\textit{``Continue straight"}} 
    \end{tabular}
    \caption{Qualitative Results for same image with different linguistic commands. Our network can successfully predict the correct navigable regions for new commands, highlighting its effectiveness in adapting to new commands flexibly.}
    \label{fig:pivot}
\end{figure}

In Figure \ref{fig:pivot}, we further scrutinize our network by fixing an image and modifying the commands to correspond to different actions. Our network is able to incorporate the changes in command and successfully reflect them in the predicted map, highlighting the network's versatility in understanding the intent of various textual commands for the same visual scene. This result showcases the controllability aspect of our network, which is highly valuable for AD applications.

\begin{figure}[t!]
    \centering
    \begin{tabular}[b]{ p{0.23\textwidth} p{0.23\textwidth}}
         \includegraphics[width=\linewidth]{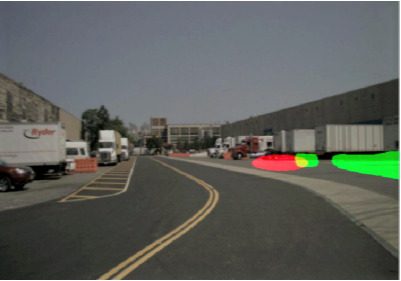}& 
        \includegraphics[width=\linewidth]{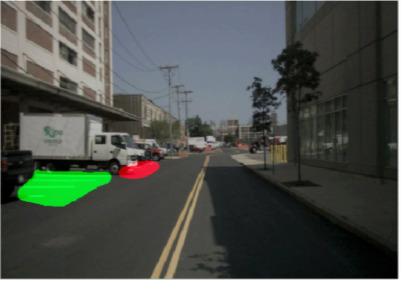}\\ 
         \centering{\scriptsize\textit{``Pull over near the first trailer on the right"}} &  
         \centering{\scriptsize\textit{``Park next to the white truck on the left side of the road"}}\\
    \end{tabular}
    \caption{Qualitative Results for Failure Cases. Even though the network fails to identify correct regions, it predicts a reasonable region near the referred object without knowing the parking rules.}
    \label{fig:failure}
\end{figure}

Some failure cases of our approach are shown in Figure \ref{fig:failure}. The results suggest that our model is able to locate the ``trailer" (in the first image) and the ``white truck" (in the second image). However, it fails to predict the navigable regions accurately. Looking closely, in the first case, the model is also able to understand the sub-phrases ``left side of the road" and ``next to white truck"; however, it predicts a place that is not appropriate for parking. The results clearly indicate the difficulty in RNR, even after correctly grounding the referred object.



\section{Navigation and Planning}

\begin{figure*}[t]
    \centering
    \begin{tabular}[b]{ p{0.28\textwidth} p{0.28\textwidth} p{0.28\textwidth}}
         \centering{\scriptsize \textit{``Turn in the direction that man is pointing to."}} &  
         \centering{\scriptsize \textit{``Park across from the white truck on the left"}} &
        \multicolumn{1}{c}{\scriptsize \textit{``Turn right before the first car on the left"}}
         \\
        \includegraphics[width=\linewidth]{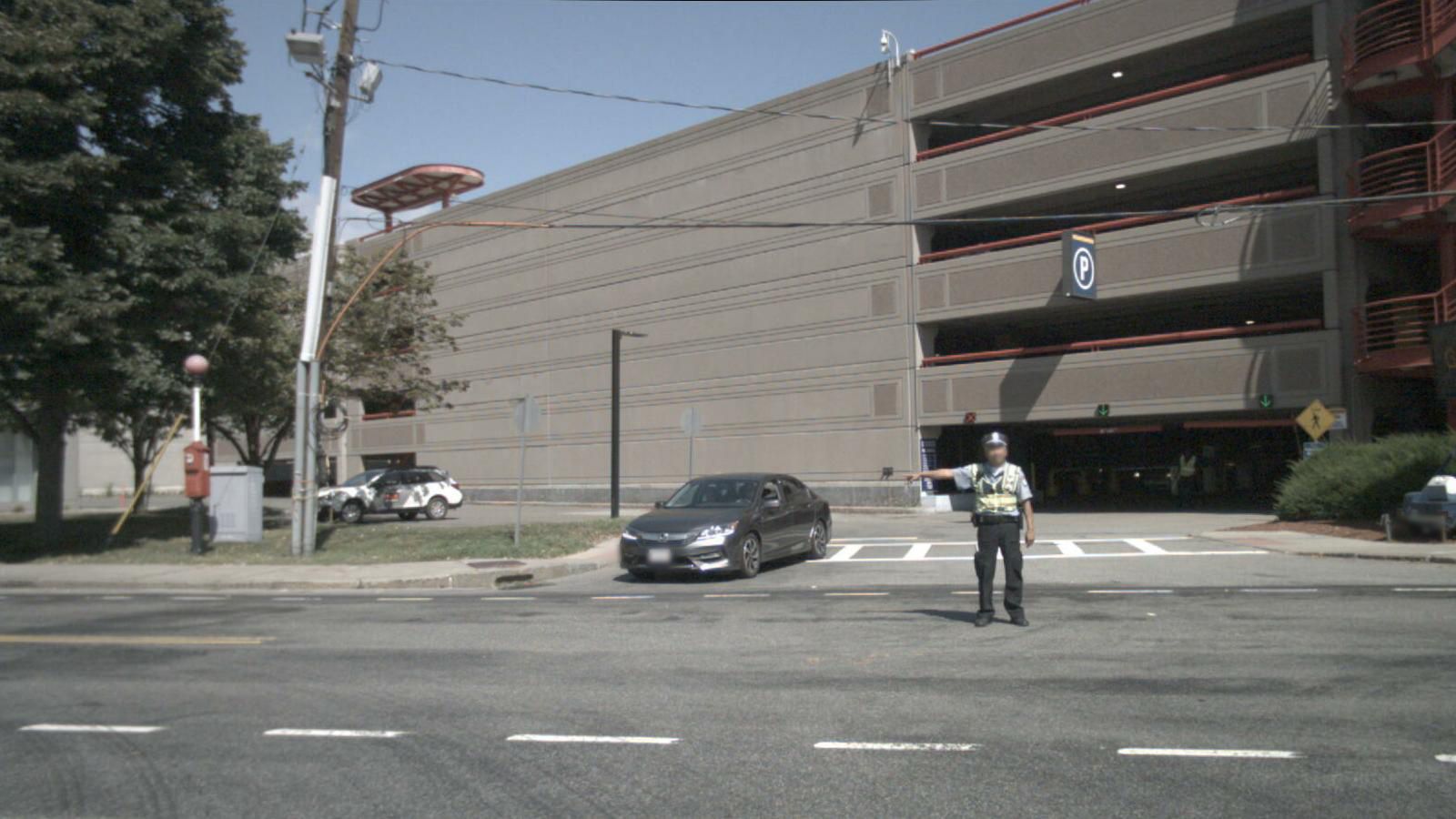}& 
        \includegraphics[width=\linewidth]{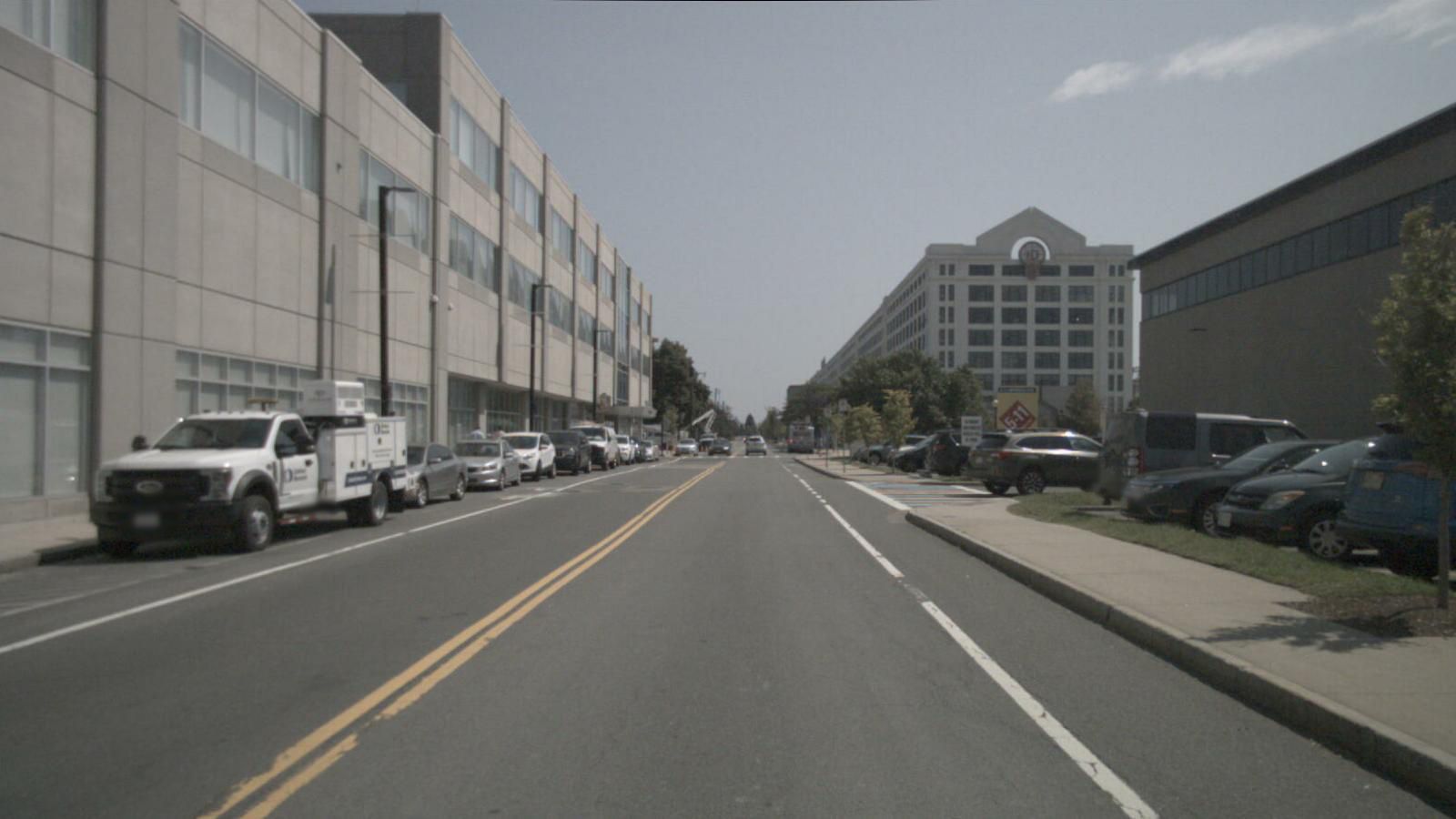}&
        \includegraphics[width=\linewidth]{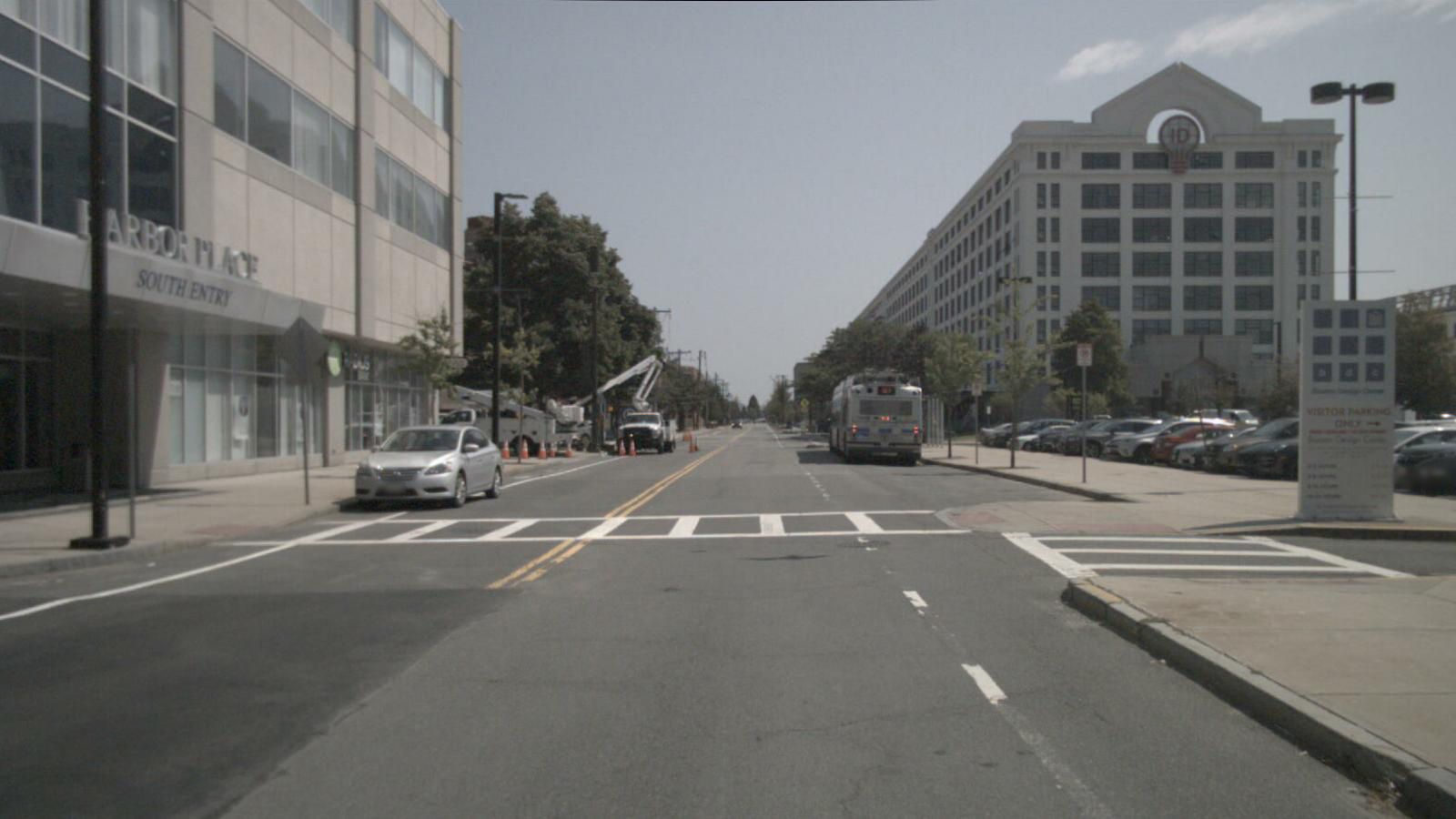} \\ 
        \includegraphics[width=\linewidth]{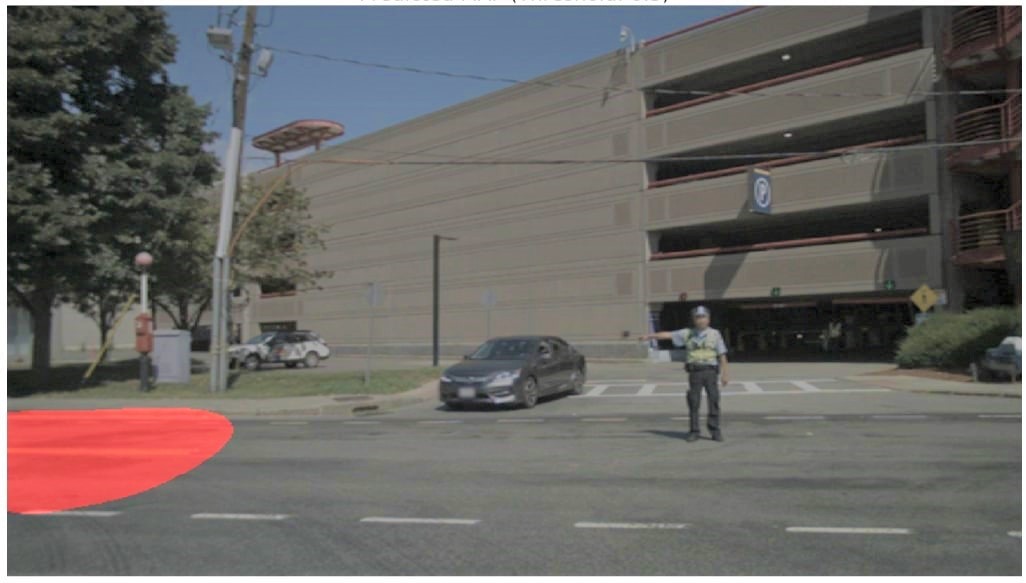}& 
        \includegraphics[width=\linewidth]{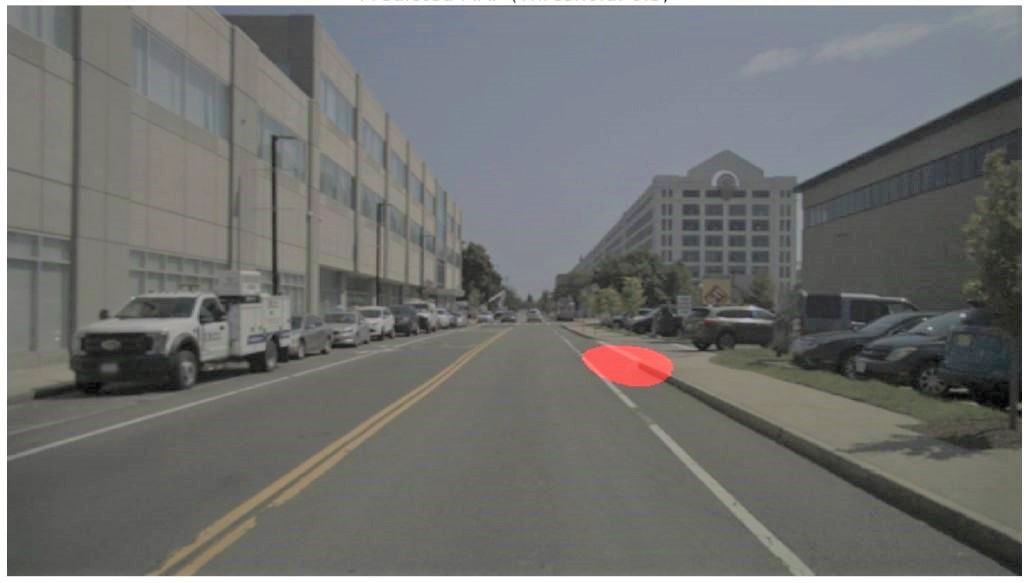}&
        \includegraphics[width=\linewidth]{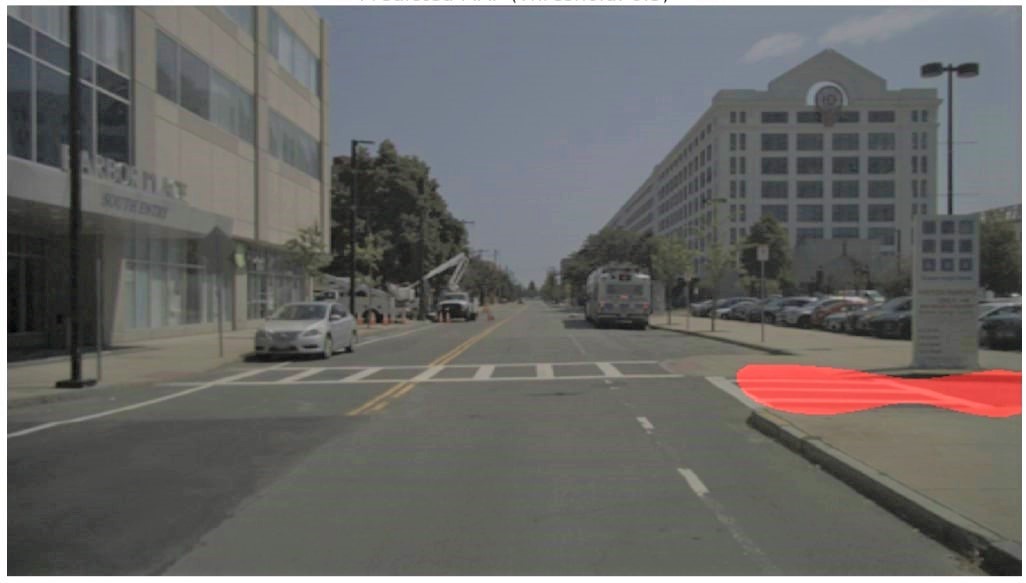} \\
         \includegraphics[width=\linewidth]{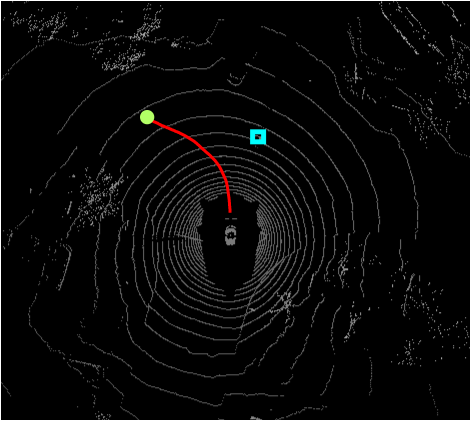}& 
        \includegraphics[width=\linewidth]{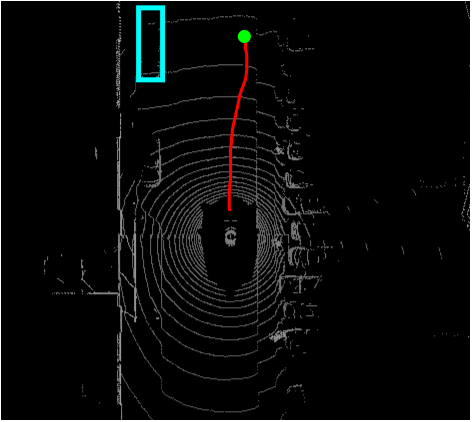}&
        \includegraphics[width=\linewidth]{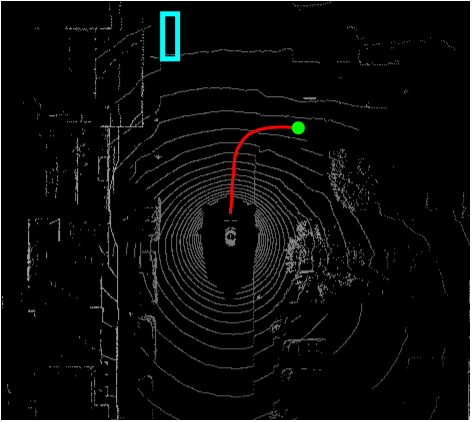} \\
    \end{tabular}
    \caption{The first row corresponds to the original image and command pairs. The second row corresponds to the predicted segmentation masks (in red) overlaid onto the images. The third row shows a feasible sample trajectory to the centre point in the predicted navigable region as a goal point}
    \label{fig:planning}
\end{figure*}

We show a downstream application wherein the navigable region output by the network is made use of by a planner to navigate to the centre of the region. While there are many potential ways of interpreting the navigable region by a downstream task, for example, one could use this as an input to a waypoint prediction network similar to \cite{song2020pip}, in this effort, we proceed with the straightforward interpretation of choosing the region centre as the goal location. 

First, we extract the ground plane from the LiDAR scan. Then we use LiDAR camera calibration to project the pixels corresponding to the grounded area in the image to the ground plane in the LiDAR scan. Finally, we use an RRT based sampling algorithm to construct a path to the point in 3D corresponding to the centre pixel of the region. 
This results in executable trajectories that appear visibly acceptable, as shown in the planned trajectories of Figure \ref{fig:planning} for a few samples from our dataset.
More involved integration to an AD application is a natural extension of this effort which will be tackled in future work.

\section{Conclusion}

This paper introduced the novel task of Referring Navigable Regions (RNR) based on linguistic commands to provide navigational guidance to autonomous vehicles. We proposed the Talk2Car-RegSeg dataset, which incorporates binary masks for regions on the road as navigational guidance for linguistic commands. This dataset is the first of its kind to enable control of autonomous vehicle's navigation based on linguistic commands. Furthermore, we propose a novel transformer-based model and present thorough experiments and ablation studies to demonstrate the efficacy of our approach. Through a downstream planner, we showed how RNR task is apt for autonomous driving applications like trajectory planning compared to the RIS task. This is the first such work which has proposed RNR and showcased its direct relevance to AD applications. In this work, we focused on single frames for grounding; future work should focus on grounding at the video-level, as it is a more realistic setting for commands with temporal constraints.


\section*{Acknowledgement}
This work was supported in part by Qualcomm Innovation Fellowship (QIF 2020) from Qualcomm Technologies, Inc.

\bibliographystyle{ieeetr}
\bibliography{iros_references}

\end{document}